\documentclass{article}
\usepackage{PRIMEarxiv}
\usepackage{hyperref}       
\usepackage[utf8]{inputenc} 
\usepackage[T1]{fontenc}    
\usepackage{booktabs}       
\usepackage{nicefrac}       
\usepackage{fancyhdr}       
\usepackage{listings}

\title{An NLP Crosswalk Between the Common Core State Standards and NAEP Item Specifications
\thanks {For reasons explained in this paper, I use edited versions of the Common Core State Standards and NAEP Item Specifications. The edited versions, which appear in the Appendices, should not be used to represent the corresponding verbatim material.} } 

\author{
  Gregory Camilli \\
  Rutgers University \\
  New Brunswick\\
  \texttt{greg.camilli@gse.rutgers.edu} \\
}

\begin{document}
\large
\maketitle

\begin{abstract} Natural language processing (NLP) is rapidly developing for applications in educational assessment. In this paper, I describe an NLP-based procedure that can be used to support subject matter experts in establishing a crosswalk between item specifications and content standards. This paper extends recent work by proposing and demonstrating the use of multivariate similarity based on embedding vectors for sentences or texts. In particular, a hybrid regression procedure is demonstrated for establishing the match of each content standard to multiple item specifications. The procedure is used to evaluate the match of the Common Core State Standards (CCSS) for mathematics at grade 4 to the corresponding item specifications for the 2026 National Assessment of Educational Progress (NAEP). 
\end{abstract}

\section {Introduction}
In this his paper, I propose a methodology using natural language processing (NLP) to map the relationship of a set of content standards to item specifications. I am not suggesting that NLP methods can substitute for human judgement, but rather that NLP methods can serve as tools to improve consistency and efficiency in the mapping process. The methodology is applied in a crosswalk from Common Core State Standards to 2026 item specifications for the National Assessment of Educational Progress (NAEP), both for grade 4 mathematics.\cite{NAGB2021, NGA} The NAEP item specifications are also termed item content areas, but I use the phrase "item specifications" throughout the paper.

This paper is organized as follows. First, a general introduction to test alignment is given, followed by a literature review of the few NLP applications currently available. The crosswalk application in this study is then described in more detail. Third, I describe the raw and transformed data. Fourth, the methods are introduced, including the crosswalk framework, semantic similarity, and a phased regression analysis. While I omit highly technical details, some description of “embedding vectors” (the transformed data) and their statistical properties is given. Results are then presented followed by discussion.

\section{Background}
The rationale for crosswalk studies is best understood in the broader context of test alignment. Educational content standards constitute the centerpiece of the American system of public education under federal and state policy. The rationale is that both tests and instruction should be aligned to those standards for maximum effectiveness. Forte described alignment as being of as being of "utmost importance because without alignment evidence — from across multiple sources — it would be impossible to interpret assessment scores in relation to the standards on which those assessments are meant to be based” (p. 6).\cite{Forte}  Given high quality standards and tests aligned to those standards, it is possible to monitor student outcomes and intervene, or at least increase awareness, when necessary. At scale, state educational standards are intended to shape instructional practices in public schools according to the concept of systemic reform.\cite{Smith}

\subsection {Types of Alignment}

There are at least three methods for aligning tests and standards for regular student populations.\cite{Webb,Blank,Achieve} While the Webb approach focuses solely on mapping test items to content standards, the Achieve approach also includes mapping item specifications to content standards.  Here, item specifications refer to a “framework that translates standards into directions for producing test items. They are concrete statements of what the test is meant to measure.”\cite{Leighton} The Blank approach further includes documents related to instruction in the evaluation of alignment.

Triangulation between content standards, test items, and instruction may include efforts to compare different tests or even to investigate repurposing an existing test. Thus, there may be an interest in understanding how an assessment like NAEP compares to an international assessment like Trends in International Mathematics and Science Study, or how the NAEP assessment compares to the CCSS in mathematics.\cite{Gattis,Daro} Studies designed to address such topics can be characterized as “crosswalks.“ This entails the idea of linking to an external criterion (a different test or set of standards), whereas a typical alignment study utilizes an internal criterion, namely test items are aligned with the standards from which they were developed. Internal and external alignment studies can be structurally similar though may they may have different objectives. Butterfuss and Doran used the more inclusive term “content mapping” to encompass alignment studies and crosswalk evaluations, and the term is broad enough to include mapping education materials to content standards.\cite{Butterfuss, Khan}

\subsection{Test Alignment Steps}
There are different approaches to alignment studies. Many studies share the following steps:

\begin{enumerate}
\item Convening a panel of subject matter experts (SME) who are trained in alignment methods.
\item Having panelists independently examine each item and identifying one or more standards to which the item can be classified as well as provide a cognitive complexity rating for the test item. 
\item Summarising results across panelists to describe the breadth and depth of standards coverage. 
\end{enumerate}

This process is not immune from substantive or semantic ambiguity in judging content matches, which is why panelist judgments are aggregated statistically or by consensus to demonstrate the preponderance of evidence. NLP may have potential for addressing or clarifying interpretive challenges, and for improving the workflow of SMEs. This assistance may increase the coherence, if not reliability, of the results of content mapping studies. 

\subsection{NLP Studies}
Khan et al. used transformer-based methods of semantic textual similarity to match educational materials (e.g., reading passages) to 24 unique Next Generation Science Standards (NGSS).\cite{Khan, NGSS} The goal was to develop a convenient system for identifying textual material targeted to particular NGSS standards, for example, matching passages from a high school biology textbook to their corresponding NGSS standards. Using embedding vectors (described below) for NGSS standards, they developed a procedure to rank and select educational content. Zhou and Ostrow examined  a data set composed of 822 standards and 9,836 test items in grades 3 through 8 English for Language Arts (ELA).\cite{Zhou}  In these data, SMEs had previously labeled test items (in traditional alignment studies) with their matching content standards. Zhou and Ostrow trained a transformer model to predict the known alignment of the test items. They found that that the correct standard was predicted with only 23-29\% accuracy. However, the average cosine similarity (described below) between embeddings vectors for a test item and its SME-determined standard was relatively high. 

Butterfuss and Doran used embedding vectors to obtain cosine similarity between the standards for 30 states and NAEP physical science “statements” and for grades 4, 8, and 12.\cite{NAGB2019} The content statements contained key science principles for NAEP assessment (NAEP does not characterize these statement as standards). Across grades, the statements covered Earth and Space Science, Life Science, and Physical Science. State content standards were crosswalked to the NAEP science statements. A series of studies carried out in advance by SMEs had matched each state standard to its corresponding NAEP statement. The goal was to determine how useful NLP methods were for correctly assigning a state standard to its true NAEP statement without reference to the SME classification decisions.
\newpage
Butterfuss and Doran showed than NLP-assisted content matching with experts (SME) can be accomplished by
\begin{enumerate}
\item Selecting a state standard.
\item Obtaining the EV similarities of this standard to each NAEP statement.
\item Sorting the similarities from highest to lowest.
\item Having SMEs work down the sorted statement lists qualitatively until an optimal NAEP match is found. 
\end{enumerate}

SMEs did not assume that the highest similarity corresponded to the best standard-statement match. Rather, the sorted list was examined to make evaluation more efficient. In fact, the highest cosine similarity did not accurately correspond to the SME-determined classifications (about 55\%), but the "true" match fell into the top five similarities about 95\% of the time. If used in a traditional alignment study, these results suggest the procedure may substantially reduce the time and resources needed for content mapping. 

\subsection{Crosswalk: Common Core Standards to NAEP Item Specifications}
Some crosswalk studies have suggested NAEP and CCSS are at least partially aligned. In particular, Daro et al. conducted a crosswalk study involving the Common Core standards and NAEP mathematics items from the 2015 assessment.\cite{Daro} Subject matter panelists included mathematicians, classroom teachers, mathematics supervisors, and mathematics educators. After training, the panelists independently matched each of 150 items from the NAEP 2015 mathematics assessment to one or more Common Core standards (Research Question 2). Daro et al. found that 77\% of Common Core standards at grades 3 and 4  could be linked to at least one NAEP item. 

In contrast to this work, the present study maps the CCSS to NAEP item specifications - not individual NAEP test items. This is useful for several reasons. First, the item specifications are more comparable in conceptual grain size to the CCSS than are individual NAEP test items. Thus, a higher-level perspective is provided. Second, of the 34 CCSS mathematics standards at grade 4, 13 involved fractions.\cite{Camilli}  NLP methods may provide a better understanding the joint coverage of the CCSS provided by the 7 NAEP item specifications that explicitly concern fractions. Third, a quantitative continuous measure of semantic similarity for standard-specification matches may be useful in assisting the evaluation of the \textit{degree} of substantive match by subject matter experts of standards to specifications. 

In this study. I only examine the directional match  Common Core standard to the set of NAEP item specifications. The match coefficients are also averaged by CCSS domain for interpretive purposes. Of course, it would also be possible to map each NAEP item specification onto the the set of Common Core standards, but this addresses a different question. Nonetheless, the reverse mapping may be useful in a future study.  

\section{Data}
The raw data is this study consist of the verbal representations of 49 NAEP item specifications and 34 Common Core standards.\cite{NAGB2020,NAGB2021,NGA} for mathematics at grade 4. I made two kinds of revision to the official texts of the data. First, in several cases a mathematical expression was rewritten in into a richer verbal description, or a verbal illustration was added. The edited specifications and standards are given in the Appendices (Sections B and C). Second, additional explanatory material was added from a draft document of the NAEP item specifications.\cite{NAGB2020}. Namely, I included examples and other information that was explicitly content-related. Several instances of this editing are shown below.

The Common Core standards are nested within 5 content domains containing the 34 standards: (Operations and Algebraic Thinking; Number and Operations in Base Ten; Number and Operations—Fractions; Measurement and Data; Geometry). The 49 NAEP items specification are also nested within 5 content domains (Number sense, properties, and operations; Algebra and Functions; Measurement; Data Analysis, Statistics, and Probability; Geometry). 

Cohesive chunks of text in the NLP literature are alternatively referred to as sentences, extended sentences, text, or statements. I use the term \textit{sentence} herein for simplicity, though I am not referring to actual grammatical sentences. Once collected, sentence data were then submitted to an large language model (LLM) to obtain vectorized semantic representations called embedding vectors (EV). These vectors are the transformed data central to the present study. With EV data, a number of analyses are possible if each vector is conceptualized as a variable, and each of its \textit{n} dimensional attributes is conceptualized as a case. An EV correlation or cosine measures a similarity between two variables, where each variable is either an item specification or a content standard, and each attribute is a case. More formally, standard and specification EVs are transposed to $n \times 1$ column vectors. With \textit{m} item specifications, the data matrix for the regression of a standard onto item specifications has dimension $n \times (m+1)$. With this classic structure, a number of conventional quantitative methods may be applied.

\section{Methods}
To use LLM technology in content mapping studies, items specifications and content standards first need to be transformed into numerical representations. These encoded vectors  are referred to as embeddings to suggest "meaning in context." For any given sentence or textual statement, an LLM takes into account the semantic properties of words and their positional context. In essence, an LLM is a massive network model with layers of predictors containing millions, if not billions of parameters. This technology uses a deep learning structure to discover the relevant endogenous predictors. Here, \textit{deep learning} refers to multiple layers of latent predictors that extract and reconstruct semantic regularities existing in large corpora of verbal material (e.g., Wikipedia, the news sources, journals, social media, etc.). This results in EVs providing information consistent with the generalized use of language expressions. 

Once estimated (or trained), the LLM can be applied to each word in a sentence of interest to obtain an EV of dimension length \textit{n} representing its semantic attributes, where \textit{n} can be large (e.g., 3000 in the present study). For a semantic representation of a sentence of \textit{k} words, the EVs of each word are then combined into a single vector to represent the sentence as a whole. However, the individual attributes of an EV vector are typically not examined because (1) they are derived analytically and have no substantive labels, (2) there are too many of them, and (3) the goal is to use numeric representations of text for prediction or classification, not to explain how or why it works. This is the infamous black box of LLMs.

A  number of LLMs for obtaining embedding vectors are available. I found that the model \texttt{text-embedding-3-large} to be useful; however, details on its architecture are sparse, possibly due to its proprietary nature.\cite{OpenAI} In any case, I used this model to obtain EVs with dimensional length \textit{n} = 3000 for each content standard and each item specification. Butterfuss and Doran compared the embeddings a number of LLMs and found that similarity coefficients varied negligibly.\cite{Butterfuss} However, new models are constantly being developed, and it is likely that substantial improvements will be made in the near term - months not years. At this date, how to choose a model for for a particular purpose is a more important question than which model to use. 

\subsection{Crosswalk Framework}
In a traditional alignment study, one focus is on how well a set of test items (A) represents a set of content standards (B). For this purpose, SMEs assign each content standard to one or more test items. A desirable outcome is that each test item is paired with one or more content standards and that SME-assigned standards match the test-development standards. In a crosswalk from content standards to item specifications (A), the idea is similar. If each standard in B can be paired with one or more specifications in A, then crosswalk from B to A exists. Moreover, if the quality and pattern of matches are consistent with the structure of B, then the crosswalk is balanced. This provides support for the claim that scores from a test based on A provide information about B. On the other hand, if the crosswalk is unbalanced or a substantial number of standards in B have no match in A, this suggests that scores from tests developed with specifications A are not comparable to those developed with content standards B. Above, I use the the words "support" and “suggests” because a traditional alignment study with actual test items may provide stronger conclusions. 

The investigation reported in the present study is on the crosswalk from Common Core standards to NAEP item specifications. This may shed some light on the interpretation NAEP scores relative to instruction grounded in the CCSS. However, crosswalks are also useful for demonstrating how two sets of content conceptually relate to one another. In the next section,  I review the ideas of similarity and multivariate similarity. I then extend the crosswalk framework to include continuous rather than discrete (yes it matches, no it doesn’t) measures of similarity. The fundamental notion of a crosswalk is not altered by these extensions.

\subsection{Similarity}
The similarity between any two sentences is commonly calculated as the cosine between their respective EVs. In the context of an LLM, two sentences with a higher cosine are interpreted as more similar in meaning. This is the basis of content-standard-to-item-specification matching: higher cosines indicate better matches. With the LLM used in this study, the cosine has an especially simple interpretation because during estimation, EVs are scaled with attributes having an average very near zero, and a sum-of-squares of 1.0. In practical terms, this scaling results in the cosine differing negligibly in value from the standard Pearson correlation. A better match is thus synonymous with a higher correlation; however,  a content standard may share unique variance with more than one item specification. 

The conceptualization of standards-specification match in terms of degree of similarity helps to address problems that arise due to imposition of discrete classification on a continuous scale of similarity. It also helps to formalize how multiple item specifications may contribute to coverage of a standard. Univariate correlation indicates the coverage of a content area by a single specification, but multiple correlation indicates the joint mapping from a content standard to multiple item specifications. To investigate multivariate coverage, I used a hybrid approach with two steps. First, the \textit{R} function \textit{VSURF\_thres} was applied to identify the the three predictors of a particular standard with the highest feature importance based on random forest regression trees.\cite{Genuer}. Subsequently, I used the \textit{R} function \textit{lm} with these three variables (in order) to conduct a hierarchical regression for obtaining unique variance contributions. Although the use of forward stepwise regression procedure is a well-known choice for variable selection in big data, it has received much criticism for this purpose.\cite{SmithG}

There were two main purposes of the regression analysis. The first purpose was to identify the item specifications onto which a content standard could be mapped. This is similar to the traditional process of discretely assigning a standard to one or more specifications. The second purpose was to obtain the multiple $R^2$ as a measure of similarity. In this sense, a higher $R^2$ indicates a higher quality match. The central premise of this approach is the goal of discovering the total amount of \textit{unique} coverage. Consider an scenario in which two item specifications are nearly identical. The regression procedure would prevent double-dipping in this case: if one of these specifications entered into the equation the next one would very likely not enter because it would not contribute additional unique variance. I would argue that redundant mapping does not result in increased content coverage. 

The hybrid stepwise procedure also provides an interpretative tool for understanding coverage substantively. The package of predictors must make sense - in the present case, mathematical sense - to be deemed quality coverage. The latter role is served by the subject matter experts. Also, interpreting information from mappings based on univariate similarity may more challenging to SMEs and more prone to inconsistency. However, nothing should prevent SMEs from rejecting the multiple regression results in their deliberations.

\subsection{Aggregate Similarity}
A set of statistics was obtained from the stepwise regression that describe the multivariate coverage of each CCSS domain. For this analysis, the squared multiple correlations $R^2$ were treated as weights, summed within CCSS and NAEP domains, and converted to percents. These statistics convey a more global view of coverage, though not the degree of similarity.

\section{Results: Standards to Specifications}
The present study focuses on semantic rather than substantive similarity. The key question is whether semantic similarity provides useful information about substantive similarity than can aid SMEs in content mapping work. In this section, the analysis of similarity in terms of stepwise regression is given followed by a description of semantic coverage. The link to substantive similarity can then either be established or rejected.

\begin{table}
\begin{center}
   \caption {Selection and Stepwise Results}
\begin{tabular}{@{}lrrrrrlllllc@{}}
\toprule
\multicolumn{2}{c}{CCCS} &  & \multicolumn{3}{c}{NAEP Import} &  & \multicolumn{3}{c}{Stepwise $R^2$}   &  & Step 1-3 \\ \cline{1-2} \cline{4-6} \cline{8-10}
Name          & Ref \#   &  & 1          & 2          & 3          &  & Step 1 & Step 2 & Step 3 &  & Increase \\ \midrule
4.OA.A.1      & 1        &  & 48         & 46         & 44        &  & 0.22   & 0.28   & 0.32   &  & 0.11     \\
4.OA.A.2      & 2        &  & 15         & 48         & 12         &  & 0.28   & 0.36   & 0.40   &  & 0.12     \\
4.OA.A.3      & 3        &  & 15         & 47         & 12         &  & 0.41   & 0.48   & 0.53   &  & 0.12     \\
4.OA.B.4      & 4        &  & 17         & 12         & 13         &  & 0.56   & 0.58   & 0.58   &  & 0.02     \\
4.OA.C.5      & 5        &  & 42         & 41         & 43         &  & 0.57   & 0.63   & 0.63   &  & 0.06     \\
4.NBT.A.1     & 6        &  & 1          & 3         & 14          &  & 0.37   & 0.42   & 0.46   &  & 0.09     \\
4.NBT.A.2     & 7        &  & 7          & 3          & 2          &  & 0.35   & 0.49   & 0.54   &  & 0.19     \\
4.NBT.A.3     & 8        &  & 1          & 3          & 2          &  & 0.32   & 0.37   & 0.41   &  & 0.08     \\
4.NBT.B.4     & 9        &  & 11         & 13         & 12          &  & 0.40    & 0.47   & 0.47   &  & 0.07     \\
4.NBT.B.5     & 10       &  & 12         & 13         & 24          &  & 0.53   & 0.55   & 0.63   &  & 0.10     \\
4.NBT.B.6     & 11       &  & 13         & 3         & 12         &  & 0.40    & 0.49   & 0.49   &  & 0.09     \\
4.NF.A.1      & 12       &  & 6          & 5         & 15         &  & 0.29   & 0.34   & 0.35   &  & 0.07     \\
4.NF.A.2      & 13       &  & 7          & 6          & 8         &  & 0.45   & 0.56   & 0.58   &  & 0.13     \\
4.NF.B.3.A    & 14       &  & 15         & 11          & 6         &  & 0.40    & 0.48 & 0.52   &  & 0.12     \\
4.NF.B.3.B    & 15       &  & 6          & 3          & 11         &  & 0.31   & 0.42   & 0.45   &  & 0.13     \\
4.NF.B.3.C    & 16       &  & 11         & 15          & 6         &  & 0.55   & 0.59   & 0.61   &  & 0.07     \\
4.NF.B.3.D    & 17       &  & 15         & 11          & 6         &  & 0.54   & 0.57   & 0.58   &  & 0.04     \\
4.NF.B.4.A    & 18       &  & 6          & 15        & 5          &  & 0.33   & 0.40   & 0.43   &  & 0.09      \\
4.NF.B.4.B    & 19       &  & 6          & 15         & 12         &  & 0.28   & 0.36   & 0.41   &  & 0.13     \\
4.NF.B.4.C    & 20       &  & 15         & 5          & 12         &  & 0.33   & 0.36   & 0.38   &  & 0.05     \\
4.NF.C.5      & 21       &  & 15         & 11          & 6          &  & 0.32   & 0.36   & 0.39   &  & 0.07     \\
4.NF.C.6      & 22       &  & 8          & 2          & 5         &  & 0.26   & 0.36   & 0.37   &  & 0.11     \\
4.NF.C.7      & 23       &  & 7          & 8         & 6          &  & 0.41   & 0.44   & 0.48   &  & 0.07      \\
4.MD.A.1      & 24       &  & 26         & 27         & 25          &  & 0.45   & 0.50   & 0.50   &  & 0.05     \\
4.MD.A.2      & 25       &  & 15         & 27         & 26         &  & 0.37   & 0.50   & 0.54   &  & 0.17     \\
4.MD.A.3      & 26       &  & 24         & 23         & 21         &  & 0.42   & 0.47   & 0.48   &  & 0.06     \\
4.MD.B.4      & 27       &  & 15         & 37         & 38         &  & 0.24   & 0.34   & 0.37   &  & 0.13     \\
4.MD.C.5.A    & 28       &  & 29         & 22         & 28          &  & 0.18   & 0.22   & 0.22   &  & 0.04     \\
4.MD.C.5.B    & 29       &  & 29         & 32          & 6         &  & 0.19   & 0.23   & 0.27   &  & 0.08     \\
4.MD.C.6      & 30       &  & 29         & 22         & 32          &  & 0.35   & 0.42   & 0.43   &  & 0.08     \\
4.MD.C.7      & 31       &  & 29         & 11         & 15          &  & 0.23   & 0.35   & 0.38   &  & 0.15     \\
4.G.A.1       & 32       &  & 29         & 36         & 28         &  & 0.53   & 0.61   & 0.65   &  & 0.11     \\
4.G.A.2       & 33       &  & 36         & 29         & 35        &  & 0.34   & 0.44    & 0.53   &  & 0.19     \\
4.G.A.3       & 34       &  & 31         & 29         & 34         &  & 0.24   & 0.32   & 0.37   &  & 0.13     \\ \bottomrule
\end{tabular}
\end{center}
\end{table}

\subsection{Similarity}
In Table 1, the results are given for the similarity of standard-to-specification matches. The first column contains the CCSS label for 34 content standards, and the second column a ordinal label for cross-referencing to the Appendix (Section B). Columns 3-5 give the reference label of the NAEP item specifications in their order of entry into the multiple regression. These ordinal labels also provide links to their counterparts in the Appendix (Section C). Columns 6-8 provide the $R^2$ at each step. For example, for 4.NF.A.2 (reference number 13), the specifications 7, 6, and 8 enter on Steps 1-3, respectively. At Step 1, the $r^2$ = .45 ($r$ = .67) is the highest univariate similarity while the multivariate similarity is $R^2$ = .58. The final column gives the additional variance explained by Steps 2-3 as 13\%. The full text of for 4.NF.A.2 is

\begin{quote}
Compare two fractions with different numerators and different denominators, e.g., by creating common denominators or numerators, or by comparing to a benchmark fraction such as 1/2. Recognize that comparisons are valid only when the two fractions refer to the same whole. Record the results of comparisons with the logical operators greater than, less than, or equal to, and justify the conclusions, e.g., by using a visual fraction model.  
\end{quote}

For this standard, the original text “>, =, or <“ was rewritten verbally as “greater than, equal to, or less than” to allow verbal processing of these operators. The three specifications corresponding to this standard as obtained from the stepwise regression are: 
\begin{quote}
\begin{description}
\item [7.] Order or compare whole numbers, decimals, or fractions using common denominators or benchmarks.
\item [6.] Recognize and generate simple equivalent (equal) fractions and explain why they are equivalent (e.g., by  using drawings).
\item [8.] Use benchmarks (well-known numbers used as meaningful points for comparison) for whole numbers, decimals, or fractions in contexts (e.g., 1/2  and 0.5 may be used as benchmarks for fractions and decimals between 0 and 1.00).
\end{description}
\end{quote}
This example demonstrates that NLP method can inform but not substitute for expert interpretation. Specifications 7 and 8 are clearly the best matching. However, specification 6 refers to a distinct but related skill, and some interpretation of relatedness is necessary. Determining whether this package of three specifications should be considered in evaluating standard coverage is  a substantive question, and addressing this issue directly may clarify the goals of the crosswalk evaluation.

Another example concerns the CCSS standard of equivalent fractions (reference number 12). This standard was rewritten from the original (4.NF.A.1) as:

\begin{quote}
    Explain why a fraction a/b is equivalent to a fraction (n × a)/(n × b) by using visual fraction models, with attention to how the number and size of the parts differ even though the two fractions themselves are the same size. Visual fraction models provide a graphical representation of fractions. Picture a rectangle divided into b equal columns with a columns shaded. The proportion of area shaded is a/b. Now divide each column into n equal pieces. There are now n x b total pieces of which n x a are  shaded. Because the shaded area in the rectangle hasn’t changed, a/b must be equal to (n x a)/(n x b). Use this principle to generate equivalent fractions.
\end{quote}
The three item specifications corresponding to this content standard are:
\begin{quote}
\begin{description}
\item[ 6.] Recognize and generate simple equivalent (equal) fractions and explain why they are equivalent
 (e.g., by  using drawings).
\item[ 5.] Connect across various representations for whole numbers, fractions, and decimals (e.g., number word, number symbol, visual representations). For example, an item might include representation of a number on a number line or with an area diagram.
\item[15.] Solve problems involving whole numbers and fractions with like denominators. Include items  that present contexts using a variety of addition/ subtraction problem structures (e.g., add to, take from, put together/ take apart, compare) and multiplication/ division problem structures  (e.g., equal groups, arrays, area, compare).

\end{description}
\end{quote}
The specifications 6, 5, and 15 enter on Steps 1-3. On Step 1, the \textit{$r^2$} = .29 (\textit{$r$} = .54) is the highest univariate similarity while the multivariate similarity at Step 3 is $R^2$ = .35. The incremental variance explained by Steps 2-3 is 7\%. All of these specifications are highly relevant to the standard even though the multiple $R^2$ is lower than the value obtained in the example above (.35 v. .58). In other words, a match was found, but the quality of the match was relatively lower than in the previous example.  This may have resulted from the textual revision of the standard.

This standard (4.NF.A.1) was rewritten from the original to include an example. The original text was
\begin{quote}
Explain why a fraction a/b is equivalent to a fraction (n × a)/(n × b) by using visual fraction models, with attention to how the number and size of the parts differ even though the two fractions themselves are the same size. Use this principle to recognize and generate equivalent fractions.
\end{quote}
The relationship between the edited version and NAEP specification 15 was unanticipated. This suggests including relevant examples in statements of standards may improve conceptualization of the content mapping process. 

The lowest values of $R^2$ were obtained for standards 4.MD.C.5.A and 4.MD.C.5.B (reference numbers 28 and 29). Both standards fall into the CCSS domain Measurement and Data and concern the conceptualization angle measurement in terms of circle arcs and rays. There is only one related NAEP item specification that involves angle measurement “Identify or draw angles and other geometric figures in the plane” (NAEP reference number 29), but this specification refers more to a skill than a concept. Nonetheless, specification 29 enters on the first step for both of these CCSS standards.

It is also interesting to examine the results for the Common Core standard 4.NBT.A.2 (reference number 7) because it shows the highest incremental in coverage in terms of $R^2$ from Step 2-3:

\begin{quote}
  Read and write multi-digit whole numbers using base-ten numerals, number names, and expanded form. Compare two multi-digit numbers based on meanings of the digits in each place, using the comparison operators greater than, equal to, and less than to record the results of comparisons.  
\end{quote}

The three item specifications corresponding to this content standard are:

\begin{quote}
\begin{description}
\item[7.] Order or compare whole numbers, decimals, or fractions using common denominators or benchmarks.
\item[3.] Compose or decompose whole quantities either by place value (e.g., write whole numbers in expanded notation using place value: 342 = 300 + 40 + 2 or 3 × 100 + 4 × 10 + 2 × 1) or convenience (e.g., to compute 4 × 27 decompose 27 into 25 + 2 because 4 × 25 is 100, and 4 × 2  is 8 so 4 × 27 is 108).
\item[2.]Represent numbers using base 10, number line, and other representations.
\end{description}
\end{quote}

The regression results are more difficult to interpret in this case. Many of the same skills are involved in the item specifications and the standard, but whether the specifications are focused enough to address the standard is an open question. This  matter requires expert deliberation. In fact, after reading the standard and the specifications, I was left wondering whether the standard would benefit from revision. It is clearly multidimensional, but I’m not clear which dimension is being or should be emphasized.

\subsection{Aggregate Similarity }

Semantic similarity can be used to help understand similarity coverage by domain for the Common Core standards and NAEP item specifications. In the top half of Table 2, the $R^2$ at Step 3 were treated as weights and summed by CCSS domain, and then taken as a percent of total weight across domains. As can be seen, most of the weight is allocated to Number and Operations—Fractions. This is not surprising. In fact, the percents in Table 2 are highly similar to the percent of standards within each domain (e.g., 13/34 = .38 for fractions standards). 

The more interesting analysis concerns the representation of the CCSS in the NAEP domains. For this analysis, the $R^2$ values were transformed to reflect unique variance. For example, in the first row of Table 1 the $R^2$ are .22, .30, and .32. These values were transformed to .22, .08 (30 - .22), and .02 (.32 - .30). The weights where then summed into the NAEP domains by first identifying the corresponding NAEP specification and then linking to its NAEP domain. After summing, the percent of total weight was calculated. It can be seen in the lower half of Table 2 that 84\% of CCSS representation in NAEP occurs in two domains: Number Sense, Properties, and Operations (68\%)
and Geometry (16\%). There is little representation in Measurement; Data Analysis, Statistics, and Probability; and Algebra.

I also examined how many times each NAEP item specification occurred in the importance order columns of Table 1. Each NAEP specification occurred at least once, but this finding tells only part of the story. Just 4 of 49 NAEP item specifications involving fractions (reference numbers 6, 11, 12,15) accounted for 40\% of the matches. 

\begin{table}
\begin{center}
 \caption {Percent of sum $R^2$ by Domain}  
\begin{tabular}{@{}llc@{}}
\toprule
Content  & Domain         & \multicolumn{1}{l}{Percent}               \\ \midrule
CCSS        & Operations and Algebraic Thinking              & 16\%      \\
            & Number and Operations in Base 10               & 19\%      \\
            & Number and Operations—Fractions                & 38\%      \\
            & Measurement and Data                           & 17\%      \\
            & Geometry                                       & 10\%      \\ \midrule
NAEP        & Number Sense, Properties, and Operations       & 68\%      \\
            & Measurement                                    &  9\%      \\
            & Geometry                                       & 16\%      \\
            & Data Analysis, Statistics, and Probability     &  1\%      \\
            & Algebra                                        &  7\%      \\ 
\bottomrule
\end{tabular}
\end{center}
\end{table}\textbf{}

\section{Discussion}

In terms of semantic textual similarity of Common Core standards to NAEP item specifications, the results presented show the potential for NLP methods to provide a conceptual toolbox in content mapping studies. Analyses based on semantic textual similarity may also provide useful reference points in content mapping studies carried out with subject matter experts. Overall, this may help to improve workflow by reducing classification inconsistencies or idiosyncratic preferences among SMEs. 

Several conclusions can be drawn regarding the results in Tables 1 and 2. First, the size of similarity coefficients requires interpretation - just like most correlation coefficients. As shown above, a lower correlation may correspond to a correct match, but the quality of the match may not be optimal. Second, if a specification is tersely written, it may only have a high similarity coefficient with a similarly terse standard. The value of this kind of content mapping is low. Adding examples to standards or specifications may be helpful, but in practice must be weighed against the risk of altering content substantively. Third, regression results such as those in Table 1 can be presented to an expert panel, but a key question concerns at what point in the mapping timeline it may be useful. Given suitable logistics, the NLP tool described in this paper appears to have value for crosswalk studies.

From a logistics perspective, obtaining embedding vectors is cheap. Only 4 \texttt{Python} statements were required, and several rounds of drawing EVs (experimenting with different lengths) cost less than one cent. Once obtained, the EV data are easily ported to any major statistical package if one prefers to work outside of the \texttt{Python} environment. The \texttt{R} code for stepwise regression is simple and direct, consisting mostly of reading in and formatting EV data. Code (with no guarantees) in \texttt{Python} and \texttt{R} is given in the Appendix (Section A) which can be modified for the reverse mapping of NAEP item specifications to Common Core standards. In Daro et al., individual SMEs physically assigned each item to just one standard, but not necessarily the same standard.\cite{Daro} In the present study, this is similar to assigning a given standard to multiple specifications. Conceptually the crosswalk mapping is from standards to specifications, though the reverse mapping could also be carried out to address a different question.

From a policy perspective, the NLP crosswalk suggests overlap between the NAEP items specifications and the CCSS, with the caveat that I used modified versions of the the CCSS and NAEP items specifications. Nonetheless, the crosswalk from standards to item specifications was highly unbalanced, which is an indication that caution should be exercised in attempting to draw conclusions from NAEP scores regarding instruction to the Common Core, at least for grade 4 mathematics. Moreover, 4 NAEP item specifications accounted for 40\% of the crosswalk, yet as few as 7 items (out of 150 or so) on a given NAEP assessment in grade 4 mathematics explicitly involve fractions on the NAEP assessment. Assessments based on the CCSS and the NAEP item specifications may provide scores based on substantially different constructs. 

The results of the present study are consistent with the work of Daro et al. who found  a "notable" difference in algebra as well as "divergences" in the reverse mappings (NAEP --> CCSS) of Geometry and  Data Analysis, Statistics, and Probability (p. 14).\cite{Daro} While they found some support for a reverse mapping in Geometry (CCSS --> NAEP). However, recall that the results in Table 1 for geometry standards indicate a "correct match," but one at a very low level of similarity. If the NLP results are sensitive to substantive similarity, it might be concluded that the only  domain in the CCSS reasonably similar to NAEP is Number Sense, Properties, and Operations. 

This research illustrates how NLP can be used to assist crosswalk studies and possibly other kinds of content mapping. It shows how a precise empirical description of the degree and structure of resemblance can be obtained to assist subject matter experts in classification tasks. I believe that NLP methods do not fundamentally change the cognitive dimensions of evaluation in content mapping, but they may be able to streamline and sharpen human judgment. NLP tools may also provide valuable information for crafting effective content standards and understanding their semantic limitations. 
\newpage

\bibliographystyle{unsrt}  
\bibliography{main}  

\begin{thebibliography}{10}

\bibitem{NAGB2021}
National Assessment~Governing Board.
\newblock {\em Mathematics Framework for the 2026 National Assessment of Educational Progress}.
\newblock Author, Washington, D.C., 2021.

\bibitem{NGA}
NGA Center \& CCSSO [National Governors Association~Center for Best~Practices and Council of~Chief State School~Officers].
\newblock {\em Common Core State Standards for Mathematics}.
\newblock Council of Chief State School Officers, Washington, D.C., 2010.

\bibitem{Forte}
E.~Forte.
\newblock {\em Evaluating alignment in large-scale standards-based assessment systems}.
\newblock Council of Chief State School Officers, Washington, D.C., 2017.

\bibitem{Smith}
M.~S. Smith and J.~O’Day.
\newblock Systemic school reform.
\newblock {\em Journal of Education Policy}, 5(5):233--267, 1990.

\bibitem{Webb}
N.~L. Webb.
\newblock {\em Criteria for alignment of expectations and assessments in mathematics and science education (Council of Chief State School Officers and National Institute for Science Education Research Monograph No. 6)}.
\newblock Wisconsin Center for Education Research, Madison, WI, 1997.

\bibitem{Blank}
A.C. Blank, R.K.and~Porter and J.~L. Smithson.
\newblock {\em New tools for analyzing teaching, curriculum and standards in mathematics and science}.
\newblock Council of Chief State School Officers, Washington, DC, 2001.

\bibitem{Achieve}
Inc. Achieve.
\newblock {\em An alignment analysis of Washington State’s college readiness mathematic standards with various local placement tests}.
\newblock Author, Cambridge, MA, 2006.

\bibitem{Leighton}
M.~J. Leighton, J. P. \&~Gierl.
\newblock Defining and evaluating models of cognition used in educational measurement to make inferences about examinees’ thinking processes.
\newblock {\em Educational Measurement: Issues and Practice}, 26:3--16, 2007.

\bibitem{Gattis}
K.~Gattis, Rosenberg S., T.~Neidorf, Guile~S. Zhang, J., and M.~McNeely.
\newblock {\em A Comparison of the2011 Grade 8 NAEP and TIMSS Mathematics and Science Frameworks (NCES 2013-462)}.
\newblock National Center for Education Statistics, Washington, D.C., 2013.

\bibitem{Daro}
P.~Daro, G.~B. Hughes, and F.~Stancavage.
\newblock {\em Study of the alignment of the 2015 NAEP mathematics items at grades 4 and 8 to the Common Core State Standards for Mathematics}.
\newblock National Assessment Governing Board, Washington, D.C., October 2015.

\bibitem{Butterfuss}
R.~Butterfuss and D.~Doran.
\newblock {\em An application of text embeddings to support alignment of educational content standards}.
\newblock Paper Presented at Generative Artificial Intelligence for Measurement and Education Meeting, feb 2024.

\bibitem{Khan}
S.~Khan, J.~Rosaler, J.and~Hamer, and T.~Almeida.
\newblock {\em Catalog: An educational content tagging system}.
\newblock Proceedings of the 14th international conference on educational data mining (EDM 2021), 2021.

\bibitem{NGSS}
NGSS~Lead States.
\newblock {\em Next Generation Science Standards: For states, by states}.
\newblock National Academies Press, Washington, D.C., 2013.

\bibitem{Zhou}
Z.~Zhou and K.S. Ostrow.
\newblock Transformer-based automated content-standards alignment: A pilot study.
\newblock In G.~et~al. Meiselwitz, editor, {\em HCI International 2022 - Late Breaking Papers. Interaction in New Media, Learning and Games. HCII 2022}, Lecture Notes in Computer Science. Springer, Cham., 2022.

\bibitem{NAGB2019}
National Assessment~Governing Board.
\newblock {\em Science framework for the 2019 National Assessment of Educational Progress}.
\newblock Author, Washington, D.C., 2013.

\bibitem{Camilli}
G.~Camilli.
\newblock The 2013-15 decline in naep mathematics in grade 4: What it teaches us about naep. measurement: Interdisciplinary research and perspectives.
\newblock {\em Measurement: Interdisciplinary Research and Perspectives}, 19(4):236--245, 2021.

\bibitem{NAGB2020}
National Assessment~Governing Board.
\newblock {\em Draft Assessment and Item Specifications for the 2026 NAEP Mathematics Assessment}.
\newblock Author, Washington, D.C., 2020.

\bibitem{OpenAI}
OpenAI.
\newblock {\em text-embedding-3-large [Large language model]}, 2024.

\bibitem{Genuer}
R.~Genuer, J-M. Poggi, and C.~Tuleau-Malot.
\newblock Vsurf: An r package for variable selection using random forests v 1.2.0.
\newblock {\em The R JOurnal}, 7(2):19--33, 2015.

\bibitem{SmithG}
G.~Smith.
\newblock Step away from stepwise.
\newblock {\em Journal of Big Data}, 5(32):1--12, 2018.

\end{thebibliography}
\newpage
\appendix

\section{Code Appendix}
\small \begin{lstlisting}
# Python code
import numpy as np
import pandas as pd
import openai
from openai import OpenAI

# Read in CCSS and NAEP Sentences [not shown here]

client = OpenAI(api_key = "Your Key Goes Here")
NE = 3000

out_CCSS = client.embeddings.create(
model = "text-embedding-3-large",
input = Sentences_CCSS,
dimensions=NE )

out_NAEP = client.embeddings.create(
model = "text-embedding-3-large",
input = Sentences_NAEP,
dimensions=NE )

# Extract and transpose Common Core
NS = len(Sentences_CCSS)
CCSS = np.zeros((NS, NE))
for i in range(NS): CCSS[i] = out_CCSS.data[i].embedding
CCSS = np.transpose(CCSS)
df = pd.DataFrame(CCSS)
df.to_csv('C:/Users/gregc/OneDrive/Desktop/CCSSM4.csv',index=False,header=False)

# Extract and transpose NAEP
NS = len(Sentences_NAEP)
NAEP = np.zeros((NS, NE))
for i in range(NS): NAEP[i] = out_NAEP.data[i].embedding
NAEP = np.transpose(NAEP)
df = pd.DataFrame(NAEP)
df.to_csv('C:/Users/gregc/OneDrive/Desktop/NAEPM4.csv',index=False,header=False)

# R code for easy correlation and regression analysis
library (leaps)
library(VSURF)
# Read in and preprocess data. Common Core & NAEP 
CCSS 	<- read.csv(file="C:/Users/gregc/OneDrive/Desktop/CCSSM4.csv", header=FALSE)
NAEP    <- read.csv(file="C:/Users/gregc/OneDrive/Desktop/NAEPM4.csv", header=FALSE)
colnames(CCSS)  <- paste("CCSS",1:34,sep="")
colnames(NAEP)  <-paste("NAEP",1:49,sep="")
# zero order correlations
CORR            <- cor(CCSS,NAEP)  

# Combine CCSS and NAEP data and create a copy
ALLD	<- data.frame(cbind(CCSS,NAEP))
ALLD2	<- ALLD 

# Random forest regression. X holds the first three predictors in terms of RF importance
X 	<- NULL
for (i in 1:34) {
DATA 	          <- ALLD2[,c(i,35:83)]
colnames(DATA)[1] <- "DV"
reg = VSURF_thres(data=DATA, DV ~ 0 + ., parallel = TRUE, mtry=6, ncores = 5)
X = rbind(X, reg$varselect.thres[1:3])
}

# Linear hiearchical regression on 3 most important variables
Y 	<- NULL
for (i in 1:34) {
    SEL	<- as.numeric( X[i,]    + 34 )
    D 	<- ALLD2[,c(i,SEL)]
    R1 	<-summary(  lm(D[,1] ~ 0 + D[,2])                    )$r.squared
    R2	<- summary (lm(D[,1] ~ 0 + D[,2] + D[,3])	        )$r.squared
    R3	<- summary( lm(D[,1] ~ 0 + D[,2] + D[,3] + D[,4])	)$r.squared
    out <-  c(R1,R2,R3)
    Y = rbind(Y, out)
}

# Second column of A contains CCSS domain labels
A           <- data.frame(matrix(0, 34, 3))
colnames(A) <- c("CCS", "Domain", "R2")

A[,1]   <- 1:34
A[,2]   <- c(1,1,1,1,1,2,2,2,2,2,2,3,3,3,3,3,3,3,3,3,3,3,3,3,4,4,4,4,4,4,4,5,5,5)
A[,3]	<- Y[,3]

COV     <- c(5,6,13,7,3)/34 # Proportion of CCSS in domains

# Calculate sum R square by domain
NUM <- NULL
for (i in 1:5) NUM <- c(  NUM,sum( subset(A, Domain==i)[,3])  ) 
TOT <- sum(NUM)
 
# Calculate percents and average
AVE <-	NUM/c(5,6,13,7,3)
PER <- 	NUM /TOT

# Prepare CCSS to NAEP domain table. Read in Table 1
Table1	 <- data.frame(  read.csv(file="C:/Users/gregc/OneDrive/Desktop/ContentMapping/Table1.csv", header=FALSE)  )
colnames(Table1) <- c("Ref", "Spec1", "Spec2", "Spec3", "R1", "R2", "R3")

# Compute unique R2
Table1$R3   <- Table1$R3 - Table1$R2
Table1$R2   <- Table1$R2 - Table1$R1

# NAEP specification-domain link
LINK        <- matrix(0,49,2)
LINK[,1]    <- 1:49
LINK[,2]    <- c(rep(1,18), rep(2,9), rep(3,9),rep(4,4),rep(5,9))

# Sum unique R2 by NAEP domain
DSUM	   <- c(0,0,0,0,0)
for (i in 1:34) {
    R <- Table1[i,]
    D <- LINK[R$Spec1,2]
        DSUM[D] <- DSUM[D] + R$R1
    D <- LINK[R$Spec2,2]
        DSUM[D] <- DSUM[D] + R$R2
    D <- LINK[R$Spec3,2]
        DSUM[D] <- DSUM[D] + R$R3
	}
DSUM/sum(DSUM)


\end{lstlisting}
\textbf{}

\newpage

\section{Modified Common Core State Standards}
\small \begin{enumerate}

\item	"Interpret a multiplication equation as a comparison, e.g., interpret 35 = 5 × 7 as a statement that 35 is 5 times as many as 7, and 35 is 7 times as many as 5. Represent verbal statements of multiplicative comparisons as multiplication equations. ",
\item	"Multiplicative Comparison: This refers to situations where you are comparing quantities by multiplication, such as ‘twice as many,’ ‘three times as much,’ or ‘half the size.’ Additive Comparison: This involves situations where you're comparing quantities by addition or subtraction, like 5 more than or 3 less than. To solve these types of problems, you may need to use drawings (visual representations) and equations with a symbol for the unknown number (often represented by a letter like x) to represent the problem. For example, if you are told that one number is twice another number, you could represent the unknown number as x and write an equation like 2x = the other number. You would then use multiplication or division to solve the equation and find the value of the unknown number. The statement emphasizes the importance of distinguishing between multiplicative comparison and additive when solving word problems.",
\item	"Solve multistep word problems posed with whole numbers and having whole-number answers using addition, subtractions, multiplication, and division, including problems in which remainders must be interpreted. Represent these problems using equations with a letter standing for the unknown quantity. Assess the reasonableness of answers using mental computation and estimation strategies including rounding. ",
\item	"Find all factor pairs for a whole number in between 1 and 100. Recognize that a whole number is a multiple of each of its factors. Determine whether a given whole number between 1 and 100 is a multiple of a given one-digit number. Determine whether a given whole number between 1 and 100 is prime or composite. ",
\item	"Generate a number or shape pattern that follows a given rule. Identify apparent features of the pattern that were not explicit in the rule itself. For example, given the rule add 3 and the starting number 1, generate terms in the resulting sequence and observe that the terms appear to alternate between odd and even numbers. Explain informally why the numbers will continue to alternate in this way. ",
\item	"Recognize that in a multi-digit whole number, a digit in one place represents ten times what it represents in the place to its right. For example, Place value refers to the value of a digit depending on its place in a number. For example, in the number 700, the digit 7 is in the hundreds place, representing 7 hundreds. The digit 0 is in the tens and ones place, representing no tens or ones. Division is a mathematical operation used to distribute a quantity into equal parts or groups. When you divide one number by another, you're essentially asking how many times the divisor fits into the dividend. So, when it says Recognize that 700 / 70 = 10 by applying concepts of place value and division, it's asking you to understand that when you divide 700 by 70, Each digit in the dividend (700) is ten times the value of the corresponding digit in the divisor (70) because of place value. Therefore, 700 divided by 70 results in 10 because 700 is ten times the value of 70. ",
\item	"Read and write multi-digit whole numbers using base-ten numerals, number names, and expanded form. Compare two multi-digit numbers based on meanings of the digits in each place, using the comparison operators greater than, equal to, and less than to record the results of comparisons. ",
\item	"Use place value understanding to round multi-digit whole numbers to any place. ",
\item	"Fluently add and subtract multi-digit whole numbers using the standard algorithm. ",
\item	"Multiply a whole number of up to four digits by a one-digit whole number, and multiply two two-digit numbers, using strategies based on place value and the properties of operations. Illustrate and explain the calculation by using equations, rectangular arrays, and/or area models. ",
\item	"Find whole-number quotients and remainders with up to four-digit dividends and one-digit divisors, using strategies based on place value, the properties of operations, and/or the relationship between multiplication and division. Illustrate and explain the calculation by using equations, rectangular arrays, and/or area models. ",
\item	"Explain why a fraction a/b is equivalent to a fraction (n × a)/(n × b) by using visual fraction models, with attention to how the number and size of the parts differ even though the two fractions themselves are the same size. Visual fraction models provide a graphical representation of fractions. Picture a rectangle divided into b equal columns with a columns shaded. The proportion of area shaded is a/b. Now divide each column into n equal pieces. There are now n x b total pieces of which n x a are  shaded. Because the shaded area in the rectangle hasn’t changed, a/b must be equal to (n x a)/(n x b). Use this principle to generate equivalent fractions.",
\item	"Compare two fractions with different numerators and different denominators, e.g., by creating common denominators or numerators, or by comparing to a benchmark fraction such as 1/2. Recognize that comparisons are valid only when the two fractions refer to the same whole. Record the results of comparisons with the logical operators greater than, equal to, or less than, and justify the conclusions, e.g., by using a visual fraction model. ",
\item	"Understand addition and subtraction of fractions as joining and separating parts referring to the same whole. ",
\item	"Decompose a fraction into a sum of fractions with the same denominator in more than one way, recording each decomposition by an equation. Justify decompositions, e.g., by using a visual fraction model. Examples: 3/8 = 1/8 + 1/8 + 1/8 ; 3/8 = 1/8 + 2/8 ; 2 + 1/8 = 1 + 1 + 1/8 = 8/8 + 8/8 + 1/8 .",
\item	"Add and subtract mixed numbers with like denominators, e.g., by replacing each mixed number with an equivalent fraction, and/or by using properties of operations and the relationship between addition and subtraction. ",
\item	"Solve word problems involving addition and subtraction of fractions referring to the same whole and having like denominators, e.g., by using visual fraction models and equations to represent the problem. ",
\item	"Understand a fraction a/b as a multiple of 1/b. For example, use a visual fraction model to represent 5/4 as the product 5 × (1/4), recording the conclusion by the equation 5/4 = 5 × (1/4). ",
\item	"Understand a multiple of a/b as a multiple of 1/b, and use this understanding to multiply a fraction by a whole number. For example, use a visual fraction model to express 3 × (2/5) as 6 × (1/5), recognizing this product as 6/5. In general, n × (a/b) = (n × a)/b. ",
\item	"Solve word problems involving multiplication of a fraction by a whole number, e.g., by using visual fraction models and equations to represent the problem. For example, if each person at a party will eat 3/8 pound of roast beef, and there will be 5 people at the party, how many pounds of roast beef will be needed? Between what two whole numbers does your answer lie? ",
\item	"Express a fraction with denominator 10 as an equivalent fraction with denominator 100, and use this technique to add two fractions with respective denominators 10 and 100. For example, express 3/10 as 30/100, and add 3/10 + 4/100 = 34/100. ",
\item	"Use decimal notation for fractions with denominators 10 or 100. For example, rewrite .62 as 62/100; describe a length as .62 meter; locate .62 on a number line diagram. ",
\item	"Compare two decimals to hundredths by reasoning about their size. Recognize that comparisons are valid only when the two decimals refer to the same whole. Record the results of comparisons with the symbols greater than, equal to, or less than, and justify the conclusions, e.g., by using a visual model. ",
\item	"Know relative sizes of measurement units within one system of units including kilometers, meters, centimeters; kilograms, grams; pounds, ounces; liters, milliliters; hours, minutes, seconds. Within a single system of measurement, express measurements in a larger unit in terms of a smaller unit. Record measurement equivalents in a two-column table. For example, know that 1 foot = 12 x  1 inch. Express the length of a 4 foot snake as 48 inches. Generate a conversion table for feet and inches listing the number pairs (1, 12), (2, 24), (3, 36), (4,48).",
\item	"Use the four operations to solve word problems involving distances, intervals of time, liquid volumes, masses of objects, and money, including problems involving simple fractions or decimals, and problems that require expressing measurements given in a larger unit in terms of a smaller unit. Represent measurement quantities using diagrams such as number line diagrams that feature a measurement scale. ",
\item	"Apply the area and perimeter formulas for rectangles in real world and mathematical problems. For example, find the width of a rectangular room given the area of the flooring and the length, by viewing the area formula as a multiplication equation with an unknown factor. ",
\item	" Make a line plot to display a data set of measurements in fractions of a unit (1/2, 1/4, 1/8). Solve problems involving addition and subtraction of fractions by using information presented in line plots. For example, from a line plot find and interpret the difference in length between the longest and shortest specimens in an insect collection. ",
\item	"Recognize and understand that An angle is measured with reference to a circle with its center at the common endpoint of the rays, by considering the fraction of the circular arc between the points where the two rays intersect the circle. An angle that turns through 1/360 of a circle is called a one-degree angle, and can be used to measure angles.",
\item	"Recognize and Understand that an angle that turns through n 1-degree 
degree angles is said to have an angle measure of n degrees. ",
\item	"Measure angles in whole-number degrees using a protractor. Sketch angles of specified measure. ",
\item	"Recognize angle measure as additive. When an angle is decomposed into non-overlapping parts, the angle measure of the whole is the sum of the angle measures of the parts. Solve addition and subtraction problems to find unknown angles on a diagram in real world and mathematical problems, e.g., by using an equation with a symbol for the unknown angle measure. ",
\item	"Draw points, lines, line segments, rays, angles (right, acute, obtuse), and perpendicular and parallel lines. Identify these in two-dimensional figures. ",
\item	"Classify two-dimensional figures based on the presence or absence of parallel or perpendicular lines, or the presence or absence of angles of a specified size. Recognize right triangles as a category, and identify right triangles. ",
\item	"Recognize a line of symmetry for a two-dimensional figure as a line across the figure such that the figure can be folded along the line into matching parts. Identify line-symmetric figures and draw lines of symmetry. "

\end{enumerate}\textbf{}

\newpage 
\section{Modified NAEP Item Specifications}
\small \begin{enumerate}

\item "Identify place value and actual value of digits in whole numbers, and think flexibly about place value notions (e.g., there are 2 hundreds in 253, there are 25 tens in 253, there are 253 ones in 253). ",
\item "Represent numbers using base 10, number line, and other representations. ", 
\item "Compose or decompose whole quantities either by place value (e.g., write whole numbers in expanded notation using place value: 342 = 300 + 40 + 2 or 3 × 100 + 4 × 10 + 2 × 1) or convenience (e.g., to compute 4 × 27 decompose 27 into 25 + 2 because 4 × 25 is 100, and 4 × 2 is 8 so 4 × 27 is 108). ",
\item "Write or rename whole numbers. (E.g., 10= 5 + 5, 10 = 12 – 2, 10 =2 x 5.) ",
\item "Connect across various representations for whole numbers, fractions, and decimals (e.g., number word, number symbol, visual representations). For example, an item might include representation of a number on a number line or with an area diagram. ", 
\item "Recognize and generate simple equivalent (equal) fractions and explain why they are equivalent (e.g., by using drawings). ",
\item "Order or compare whole numbers, decimals, or fractions using common denominators or benchmarks. ",
\item "Use benchmarks (well-known numbers used as meaningful points for comparison) for whole numbers, decimals, or fractions in contexts (e.g., 1/2  and 0.5 may be used as benchmarks for fractions and decimals between 0 and 1.00). ",
\item "Make estimates appropriate to a given situation with whole numbers, fractions, or decimals.",
\item "Verify solutions or determine the reasonableness of results in meaningful contexts. For example, an item might require justification for a whole number response based on the context used in division involving a remainder.",
\item "Add and subtract using conventional or unconventional procedures (e.g., strategic decomposing and composing): Whole numbers, or Fractions and mixed numbers with like denominators.) ",
\item "Multiply numbers using conventional or unconventional procedures (e.g., strategic decomposing and composing): Whole numbers no larger than two digits by two digits with paper and pencil computation, or Larger whole numbers using a calculator, or Multiplying a fraction by a whole number.",
\item "Divide whole numbers: Up to three digits by one digit with paper and pencil computation, or Up to five digits by two digits with use of calculator.", 
\item "Interpret, explain, or justify whole number operations and explain the relationships between them. Emphasis should be on interpreting, explaining, or justifying subtracting a number as the inverse operation to adding a number, or dividing by a number as the inverse operation to multiplying a number. " ,
\item "Solve problems involving whole numbers and fractions with like denominators. Include items that present contexts using a variety of addition/ subtraction problem structures (e.g., add to, take from, put together/ take apart, compare) and multiplication/ division problem structures (e.g., equal groups, arrays, area, compare).",
\item "Identify odd and even numbers. Include items that involve determining whether the number of objects in a given set is even or odd. Include items that involve writing an even number as the sum of two equal addends or as a sum of twos. ",
\item "Identify factors of whole numbers. Items should involve identification of single-digit factors of whole numbers through 100. ",
\item "Apply basic properties of operations. Items should involve the commutative and associative properties of addition and multiplication, the distributive property of multiplication across addition, the identity property of addition, and multiplication by zero. ",
\item "Identify the attribute that is appropriate to measure in a given situation. ",
\item "Compare objects with respect to a given attribute, such as length, area, volume, time, or temperature. ",
\item "Estimate the size of an object with respect to a given measurement attribute (e.g., length, perimeter, or area using a grid). For example, an item might require estimating the area of an irregular shape presented on a grid.",
\item "Select or use appropriate measurement instruments such as ruler, meter stick, clock, thermometer, or other scaled instruments, including a protractor.",	
\item "Solve problems involving perimeter of plane figures. Plane figures can be polygons but cannot be circles.",
\item "Solve problems involving area of squares and rectangles. Include items that relate area to the operations of multiplication and addition, such as tiling a rectangle with whole-number side lengths and showing that the area is the same as would be found by multiplying the side lengths.",
\item "Select or use an appropriate type of unit for the attribute being measured such as length, time, or temperature. ",
\item "Solve problems involving conversions within the same measurement system such as conversions involving inches and feet or hours and minutes. ",
\item "Determine appropriate size of unit of measurement in problem situation involving such attributes as length, time, capacity, or weight. ",
\item "Identify or describe (informally) real-world objects using simple plane figures (e.g., triangles, rectangles, squares, and circles) and simple solid figures (e.g., cubes, spheres, and cylinders). ",
\item "Identify or draw angles and other geometric figures in the plane.",
\item "Describe or distinguish among attributes of 2-dimensional and 3-dimensional shapes. ",
\item  " Recognize attributes (such as shape and area) that do not change when plane figures are subdivided and rearranged. ",
\item  "Analyze or describe patterns in polygons when the number of sides increases, or the size or orientation changes. ",
\item  "Combine simple plane shapes to construct a given shape. Include items that involve combining two-dimensional shapes to construct a three-dimensional figure.  ",
\item  "Recognize 2-dimensional faces of 3-dimensional shapes.",
\item  "Describe and compare properties of simple and compound figures composed of triangles, squares, and rectangles. For example, an item might provide a rectangular prism and require identification of the faces that have the same area.  ",
\item "Describe relative positions of points and lines using the geometric ideas of parallelism or perpendicularity.",
\item  "Read or interpret a single distribution of data.",
\item  "For a given distribution of data, complete a graph. ",
\item "Answer statistical questions by estimating and computing within a single distribution of data. ",
\item  "Given a distribution of whole number data in a context, identify and explain the meaning of the greatest value, the least value, or of any clustering or grouping of data in the distribution. ",
\item  "Recognize, describe (in words or symbols), or extend simple numerical and visual patterns. Items should assess extensions of patterns in mathematically appropriate ways. For example, patterns should either be presented in ways that are transferable to a larger set or allow for multiple correct responses when not transferable to a larger set (e.g., when the first six elements of a pattern do not necessarily indicate the next six elements). Pattern types can include whole numbers or shapes. ",
\item  "Given a description, extend or fiind a missing term in a pattern or sequence. ",
\item  "Create a different representation of a pattern or sequence given a verbal description. ",
\item  "Translate between different representational forms (symbolic, numerical, verbal, or pictorial) of whole number relationships (such as from a written description to an equation or from a function table to a written description).",
\item  "Use letters and symbols to represent an unknown quantity in a simple mathematical expression.",
\item  "Express simple mathematical relationships using expressions, equations, or inequalities. ",
\item "Find the unknown(s) in a whole number sentence (e.g., in an equation or simple inequality like x + 3 > 7). Items should present equations and inequalities that involve no more than one operation in the process of determining an unknown or a set of unknowns",
\item  "Interpret the symbol = as an equivalence between two values and use this interpretation to solve problems. ",
\item  "Verify a conclusion using simple algebraic properties derived from work with numbers (e.g., commutativity, properties of 0 and 1). For example, an item might require understanding that if Sam is 3 years older than Ned, 20 years from now Sam will still be 3 years older than Ned."
\end{enumerate}\textbf{}

\end{document}